\documentclass[runningheads]{llncs}
\usepackage{graphicx}
\usepackage[table]{xcolor}
\usepackage{booktabs}
\usepackage{multirow}
\usepackage{amsmath}
\usepackage{amssymb}
\usepackage{adjustbox}
\usepackage{placeins}
\usepackage{float}
\usepackage{url}

\newcommand{\methodfig}[2]{%
  \IfFileExists{figures/#1}{%
    \includegraphics[width=\linewidth]{figures/#1}%
  }{%
    \fbox{\parbox{0.98\linewidth}{\centering\vspace{1.2cm}%
    \textbf{#2}\vspace{1.2cm}}}%
  }%
}

\begin{document}

\title{AOV-GS: Active Open-Vocabulary Semantic Mapping with 3D Gaussian Splatting}
\titlerunning{AOV-GS}

\author{Rodion Shkokov\inst{1}\textsuperscript{*} \and
Tatiana Zemskova\inst{1,2} \and Dmitry Yudin\inst{1,2}}
\authorrunning{R. Shkokov et al.}
\tocauthor{Rodion Shkokov, Tatiana Zemskova, Dmitry Yudin}
\institute{Moscow Institute of Physics and Technology (MIPT), Dolgoprudny, Russia\\
\email{r.shkokov@gmail.com} (\textsuperscript{*}corresponding author)
\and AIRI, Moscow, Russia}

\maketitle

\begin{abstract}
Robots deployed in human-centered environments need maps that support both geometric
navigation and open-vocabulary semantic queries beyond a fixed label set.
A key challenge is enriching such maps with semantic information without relying on
manually annotated segmentation masks or closed-set semantic categories.
Existing geometry-driven active mapping maximizes coverage but treats semantics as a passive
by-product of the collected trajectory, whereas active semantic mapping typically depends on
supervised semantic segmentation.
We present \textbf{AOV-GS}, an active 3D Gaussian mapping system that guides exploration
using frozen foundation models.
During coarse exploration, AOV-GS re-ranks geometrically promising candidate views according
to CLIP mask-embedding novelty computed from SAM instance segmentation, biasing exploration
toward semantically informative regions.
Frozen SAM and CLIP provide instance masks and language-aligned embeddings throughout
mapping, while a lightweight scene-specific language field is fitted offline directly to the
collected CLIP embeddings rather than manually annotated masks, enabling open-vocabulary
text-query segmentation on novel views.
In a controlled comparison using the same SAM--CLIP perception stack, AOV-GS improves average
language-field mIoU over geometry-only active planning by 1.3 percentage points on Replica
(34.6\% on eight scenes) and by 1.8 points on Matterport3D (23.1\% on five scenes).
With CorrCLIP, AOV-GS reaches 35.6\% on Replica, 2.0 points above the matched
Geometry-based + CorrCLIP baseline.
Novel-view PSNR remains comparable to the geometry-only baseline on both datasets while depth
L1 improves on Replica.
Code is available at \url{github.com/OptimusFaber/AOV-GS}.
\end{abstract}

\section{Introduction}

Mobile robots operating in human-centered environments need spatial maps of the environment
that support both navigation and task-level semantic interaction.
In practice this means a map should render high-quality RGB-D views \emph{and} support queries
such as \emph{find the chair} or \emph{locate the table near the window}.
These objectives are coupled through exploration: the viewpoints collected during mapping
determine geometric completeness, object visibility, and the evidence available for
downstream open-vocabulary retrieval.

Recent 3D Gaussian Splatting (3DGS) SLAM systems~\cite{kerbl20233dgaussiansplatting,keetha2024splatam}
achieve efficient high-fidelity reconstruction suitable for online robotic mapping.
Active planners such as ActiveGS~\cite{jin2025activegs}, ActiveSplat~\cite{li2025activesplat},
and ActiveGAMER~\cite{chen2025activegamer} improve coverage through geometry-aware
next-best-view (NBV) selection.
In parallel, foundation-model perception enables open-vocabulary semantics:
Segment Anything (SAM)~\cite{kirillov2023segment} provides prompt-free instance masks and
CLIP~\cite{radford2021learning} maps mask regions to a language-aligned embedding space.
LangSplat~\cite{qin2024langsplat} and related methods attach such features to 3D primitives
for text-query segmentation, but typically assume a \emph{fixed} observation trajectory.

Active semantic mapping closes part of this gap.
ActiveSGM~\cite{chen2025activesgm} demonstrates that semantic signals can guide exploration in a
3DGS loop, but relies on per-scene OneFormer~\cite{jain2023oneformer} fine-tuning, which limits
open-vocabulary deployment.
Passive fusion pipelines~\cite{jatavallabhula2023conceptfusion,peng2023openscene} can improve
semantic quality on pre-recorded trajectories but do not affect online data acquisition.

This paper addresses the following question:
\emph{Can a robot actively explore an unseen scene using frozen foundation models,
improve open-vocabulary segmentation through viewpoint selection,
and preserve competitive RGB-D reconstruction quality?}

We answer affirmatively with \textbf{AOV-GS} (Fig.~\ref{fig:pipeline_neuro}).
During coarse exploration (stage~0), geometric information gain is augmented with
CLIP-embedding novelty computed from SAM instance masks on RGB views rendered from the
current Gaussian map.
Semantic novelty is measured against a growing keyframe embedding bank and re-ranks the
top-$K$ geometrically strong candidates.
During refinement (stage~1+), planning reverts to geometry-only behavior as in
ActiveGS~\cite{jin2025activegs}.
After exploration, per-keyframe CLIP mask embeddings train a sparse language field on Gaussian
primitives, enabling open-vocabulary text-query segmentation on novel views.
As in ActiveSGM, candidate RGB views for semantic scoring are rendered from the current
Gaussian map; Habitat supplies the posed RGB-D observation after the selected view is executed.
The experiments nevertheless use ground-truth camera poses as oracle SLAM input.

Our contributions are:
\begin{itemize}
\item A coarse exploration NBV criterion that re-ranks the top-$K$ candidates produced by
the geometric planner according to CLIP mask-embedding novelty, steering the robot toward
semantically novel object instances while reverting to geometry-only scoring during later
refinement stages.
\item An active open-vocabulary semantic mapping pipeline combining online semantic
exploration with offline LangSplatV2 language-field construction, without per-scene
segmenter fine-tuning.
\item A two-benchmark evaluation showing average open-vocabulary gains over geometry-only
exploration while preserving comparable novel-view reconstruction quality. Under the
matched SAM--CLIP stack, AOV-GS reaches 34.6\% on Replica (+1.3~pp) and 23.1\% on MP3D
(+1.8~pp); the per-scene Replica results are heterogeneous and are analyzed explicitly.
\end{itemize}

\section{Related Work}

Active perception for neural mapping uses information gain and coverage criteria to select
informative views~\cite{jiang2023fisherrf,feng2024naruto,pan2022activenerf}.
3D Gaussian Splatting (3DGS)~\cite{kerbl20233dgaussiansplatting,keetha2024splatam} lowers
planning and rendering cost relative to Neural Radiance Fields
(NeRF)~\cite{mildenhall2021nerf}, enabling online systems such as
ActiveGS~\cite{jin2025activegs}, ActiveSplat~\cite{li2025activesplat}, and
ActiveGAMER~\cite{chen2025activegamer}; these methods optimize geometry but leave semantic
map utility to the collected trajectory.
Semantic and open-vocabulary Gaussian mapping~\cite{li2024sgsslam,siegel2025segsplat,qin2024langsplat,kerr2023lerf,cen2023saga,zhou2024feature3dgs}
attaches language-aligned features to 3D primitives for text queries, typically on fixed
trajectories; our language-field head follows LangSplatV2~\cite{li2025langsplatv2}, but AOV-GS uses semantic novelty
to \emph{choose} views during exploration rather than only processing a passive path.
Recent semantic mapping systems also include LEG-SLAM~\cite{titkov2025legslam} and
M3DMap~\cite{yudin2025m3dmap}, which integrate language-aligned or object-aware semantics
into 3D reconstruction.
Closely related but complementary is SegmATRon~\cite{zemskova2023segmatron}, an embodied
\emph{active segmentation} method that generates agent actions to maximize image
segmentation quality rather than building a persistent semantic map.
ActiveSGM~\cite{chen2025activesgm} couples geometric and semantic terms during 3DGS
exploration but requires per-scene OneFormer fine-tuning~\cite{jain2023oneformer}.
Passive open-vocabulary fusion~\cite{jatavallabhula2023conceptfusion,peng2023openscene}
improves semantics on recorded paths without steering acquisition.
AOV-GS trades closed-set semantic supervision for open-vocabulary scene representations
based on frozen SAM and CLIP.

\section{Method}

\begin{figure}[t]
    \centering
    \methodfig{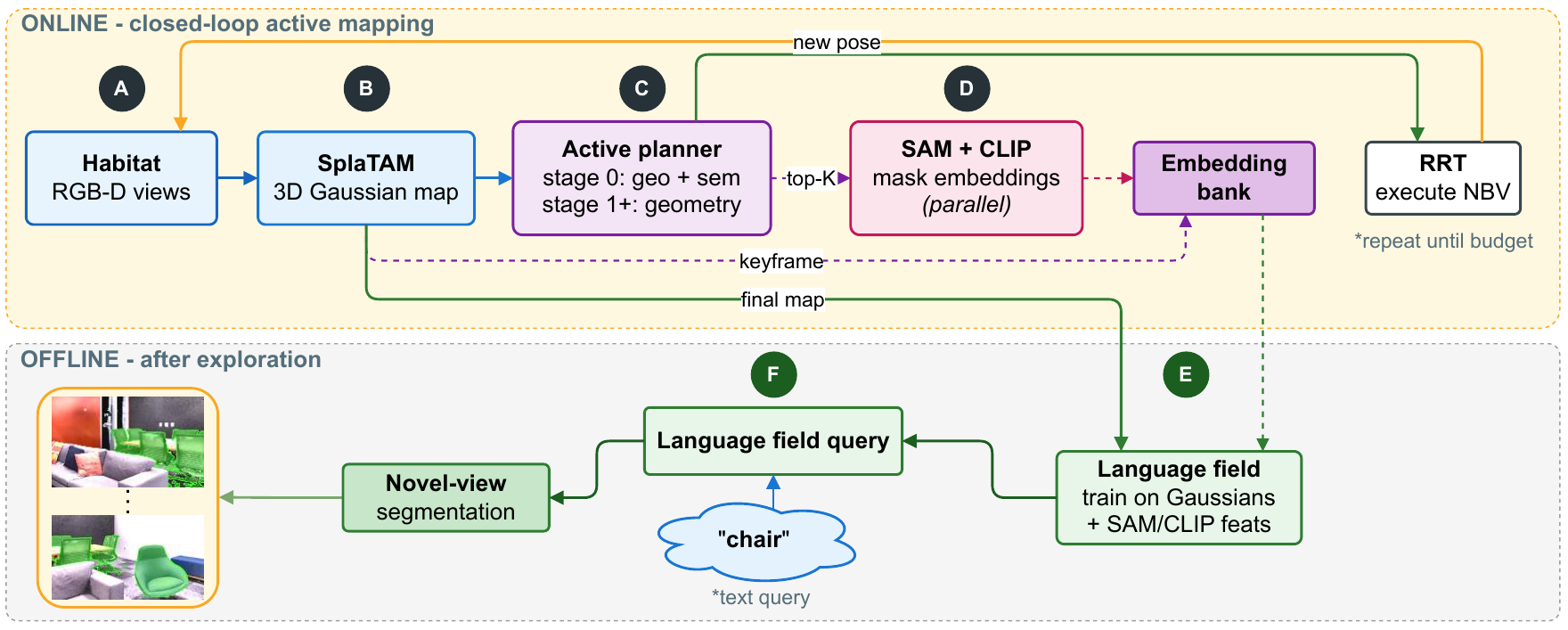}{AOV-GS pipeline overview}
    \caption{AOV-GS system overview.
    \textbf{Online loop}: posed RGB-D updates a Gaussian map; the planner selects NBVs;
    CLIP mask embeddings are extracted in parallel on keyframes.
    \textbf{Offline stage}: a language field is trained on Gaussians and queried by text.}
    \label{fig:pipeline_neuro}
\end{figure}

Figure~\ref{fig:pipeline_neuro} gives a system-level overview:
\textbf{(A--B)} posed RGB-D mapping (Sec.~\ref{sec:method-mapping});
\textbf{(C)} active NBV planner with semantic stage~0 and geometry-only stage~1+
(Sec.~\ref{sec:method-semantic}; Fig.~\ref{fig:geom_neuro});
\textbf{(D)} CLIP mask embeddings (Sec.~\ref{sec:method-embeddings}; Fig.~\ref{fig:samclip_neuro});
\textbf{(E--F)} language-field training/query (Sec.~\ref{sec:method-langfield}; Fig.~\ref{fig:lang_neuro}).

\subsection{Gaussian Mapping Backbone}
\label{sec:method-mapping}

We adopt SplaTAM~\cite{keetha2024splatam} as the reconstruction backbone with ActiveGS-style
active planning infrastructure~\cite{jin2025activegs}.
Habitat~\cite{savva2019habitat} provides synchronized RGB-D frames with ground-truth camera
poses.
In both Replica and Matterport3D experiments, these poses are used as oracle SLAM input
(Sec.~\ref{sec:experiments-setup}).
The scene is represented by isotropic 3D Gaussians optimized through differentiable rendering
with RGB and depth $L_1$ losses over silhouette pixels~\cite{keetha2024splatam,jin2025activegs}.
Following ActiveGS~\cite{jin2025activegs}, we maintain a voxel-based exploration map for collision-free candidate
pose sampling and execute selected views with an RRT-based local planner~\cite{lavalle1998rrt}.

\subsection{Coarse Exploration Criterion}
\label{sec:method-semantic}

Exploration proceeds in two stages following ActiveGAMER~\cite{chen2025activegamer}:
\textbf{stage~0} augments geometry with semantic novelty; \textbf{stage~1+} reverts to pure
geometry as in ActiveGS~\cite{jin2025activegs}.

\paragraph{Geometric score.}
For each candidate pose $v$, we render the current map and count unseen silhouette pixels
$N_v^{S}$.
Following ActiveGAMER~\cite{chen2025activegamer}, the geometric information gain is
\begin{equation}
I_v^{\mathrm{geo}}=\sigma(\log N_v^{S}),
\end{equation}
where $\sigma(\cdot)$ denotes softmax normalization over all candidates.
Let $T_x^v\in\mathbb{R}^3$ and $T_x^t\in\mathbb{R}^3$ denote, respectively, the translation
of candidate pose $v$ and the robot translation at planning step $t$.
With distance $l_v=\|T_x^v-T_x^t\|_2$, the distance-weighted score is
\begin{equation}
\label{eq:geo_weight}
w_v^{\mathrm{geo}}=\bigl(1-\sigma(l_v)\bigr)\,I_v^{\mathrm{geo}}.
\end{equation}

\paragraph{Semantic novelty.}
For the top-$K$ candidates ranked by $w_v^{\mathrm{geo}}$, the current Gaussian map renders an
RGB view at each candidate pose. We apply SAM to obtain instance masks and encode each masked region with CLIP to produce
embeddings $\{e_i\}$, $e_i\in\mathbb{R}^{512}$ (Sec.~\ref{sec:method-embeddings}).
Thus semantic scoring uses only the appearance currently represented by the map, matching the
information available to the geometry branch and the corresponding ActiveSGM setup.
Each $e_i$ is compared against keyframe bank $\mathcal{B}$.
Candidate embeddings are used only transiently to score the current candidate set and are
not inserted into $\mathcal{B}$; the bank contains embeddings extracted from RGB observations
at accepted keyframes reached by the executed trajectory.
Let $\mathrm{sim}(a,b)$ denote cosine similarity.
The semantic novelty score is
\begin{equation}
\label{eq:sem_novelty}
I_v^{\mathrm{sem}}=\mathrm{Agg}_{i}\!\left(1-\max_{b_j\in\mathcal{B}}\mathrm{sim}(e_i,b_j)\right),
\end{equation}
where $\mathrm{Agg}$ is mean pooling over masks and $b_j$ denotes the $j$-th embedding stored
in bank $\mathcal{B}$.
To avoid an empty reference set at the start of exploration, semantic scoring is skipped until
$\mathcal{B}$ contains at least one embedding (the configured threshold is
$n_{\min}^{\mathcal B}{=}1$). We use distinct notation because $\tau_q$ denotes the query
decoding threshold below.

\begin{figure}[t]
    \centering
    \methodfig{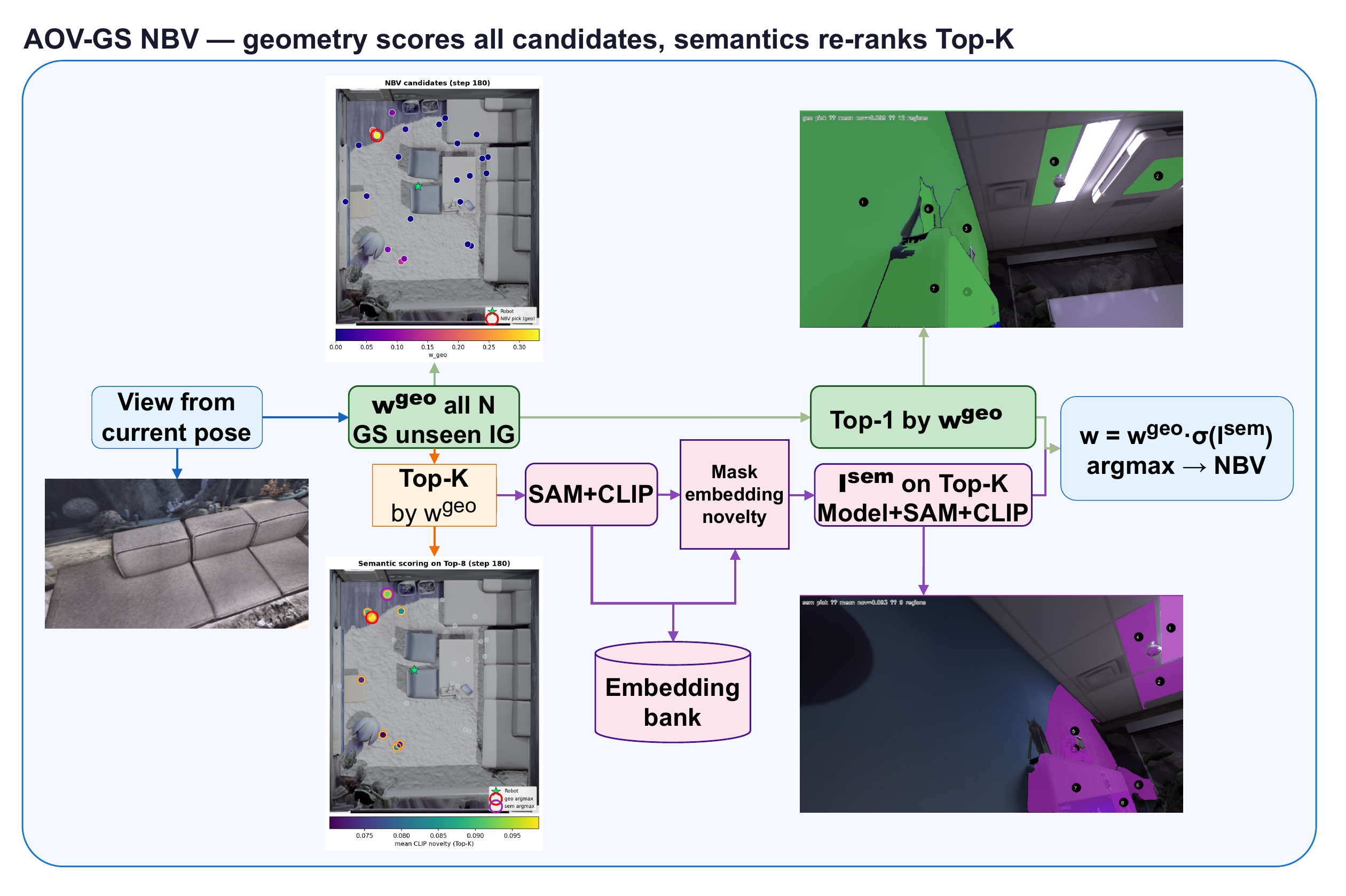}{Geometry vs.\ semantic NBV}
    \caption{Stage-0 NBV on \texttt{office0}: candidate poses, geometric scores, and
    geometry-only vs.\ semantic-aware argmax.}
    \label{fig:geom_neuro}
\end{figure}

\paragraph{Combined criterion.}
For each candidate pose $v$, the fused planner score is
\begin{equation}
w_v = w_v^{\mathrm{geo}}\cdot \sigma(I_v^{\mathrm{sem}}),
\end{equation}
where $w_v^{\mathrm{geo}}$ and $I_v^{\mathrm{sem}}$ are defined in
Eqs.~\eqref{eq:geo_weight} and~\eqref{eq:sem_novelty}; the NBV is $\arg\max_v w_v$.
Semantic novelty \emph{re-weights} geometrically strong candidates rather than replacing
geometry-driven planning: when geometric scores among top candidates are close, the product
can move the argmax toward views whose rendered masks contain embeddings far from $\mathcal{B}$
(Fig.~\ref{fig:geom_neuro}).

\subsection{Parallel CLIP Mask Embedding Extraction}
\label{sec:method-embeddings}

\begin{figure}[ht]
    \centering
    \methodfig{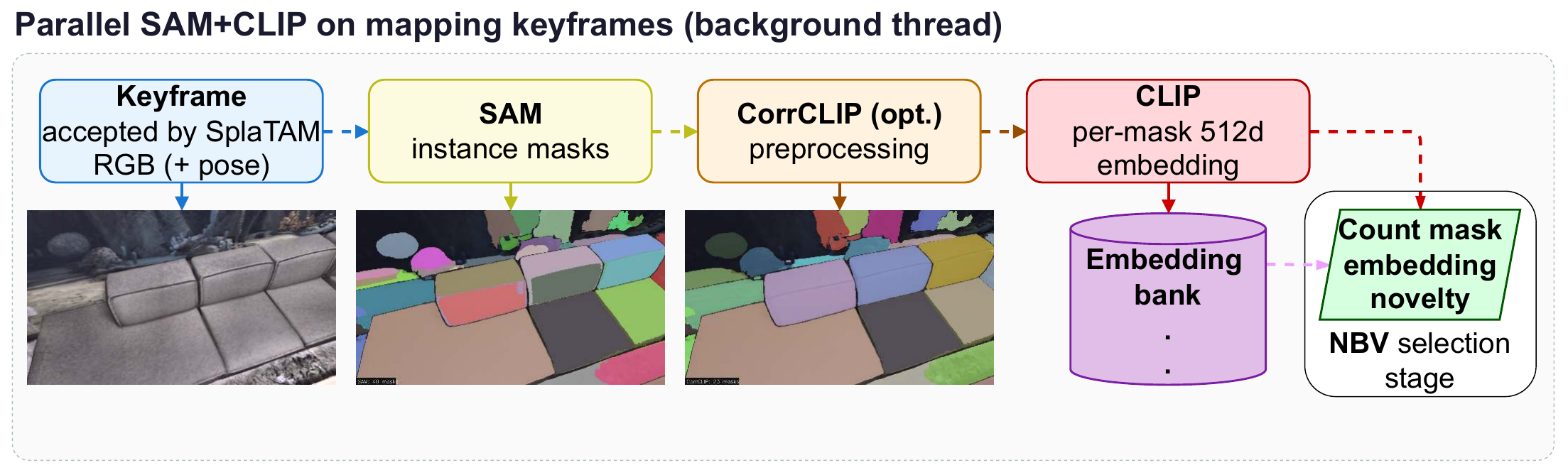}{CLIP mask embedding pipeline}
    \caption{SAM--CLIP extraction on keyframes with optional CorrCLIP post-processing;
    embeddings feed $\mathcal{B}$ and offline language-field training.}
    \label{fig:samclip_neuro}
\end{figure}

For each accepted mapping keyframe, SAM ViT-H~\cite{kirillov2023segment} segments the RGB image
into instance masks; CLIP ViT-B/16~\cite{radford2021learning} encodes each region into a
$512$-D embedding (Fig.~\ref{fig:samclip_neuro}).
We refer to this two-step process as the \emph{SAM--CLIP pipeline}.
Following LangSplatV2~\cite{li2025langsplatv2}, per-mask vectors supervise offline
language-field distillation and populate bank $\mathcal{B}$ (Eq.~\eqref{eq:sem_novelty}).
SAM and CLIP are \emph{frozen}; no per-scene fine-tuning is performed.
We optionally apply CorrCLIP-style~\cite{zhao2024corrclip} mask merging and inter-class
suppression to stabilize downstream training.
Extraction runs in a background thread and does not block the online planner.

\subsection{Language Field Construction and Query}
\label{sec:method-langfield}

\begin{figure}[ht]
    \centering
    \methodfig{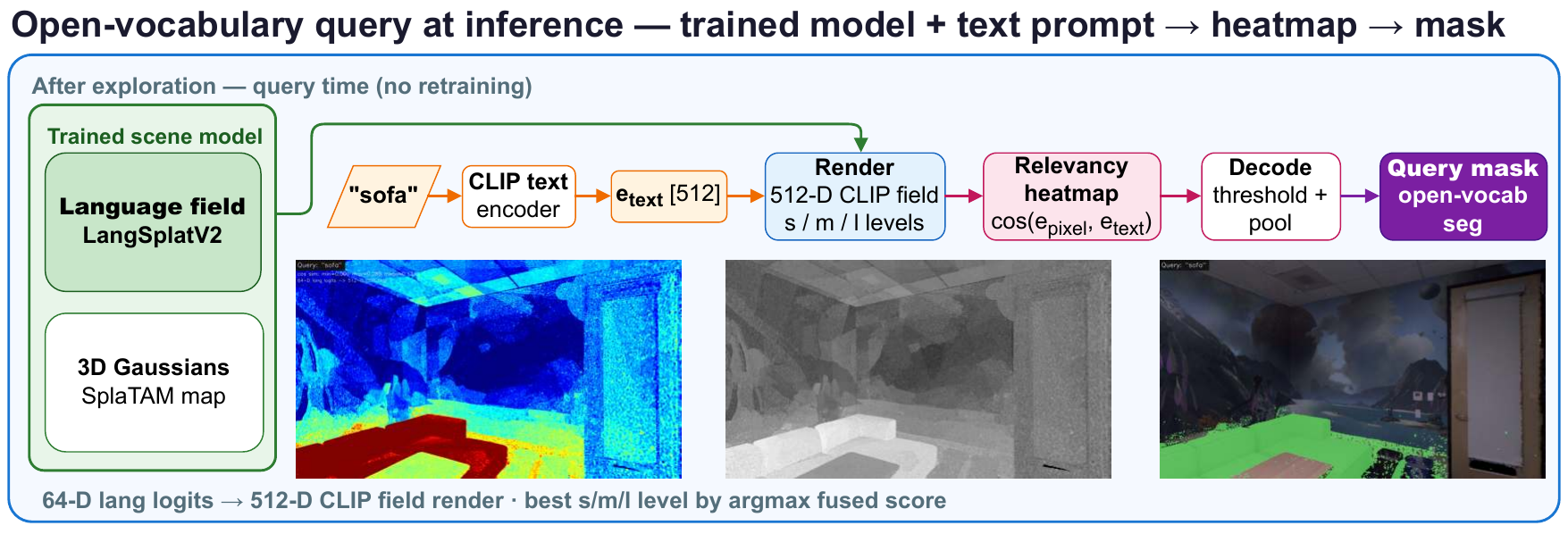}{Language-field text query}
    \caption{Open-vocabulary text query via multi-scale cosine-similarity maps and thresholding.}
    \label{fig:lang_neuro}
\end{figure}

After exploration, stored CLIP mask embeddings train a LangSplatV2 sparse language
field on the reconstructed Gaussians.
A global codebook is initialized by k-means over mask embeddings; each Gaussian receives sparse
coefficients over codebook entries.
At inference, a text prompt is encoded by CLIP and matched against multi-scale rendered
cosine-similarity maps ($s/m/l$), producing an open-vocabulary segmentation mask on a novel view
(Fig.~\ref{fig:lang_neuro}).
The language field is trained \emph{after} mapping and is decoupled from geometric refinement.

\subsection{Implementation Details}

Default online settings use top-$K{=}8$ semantic candidates in stage~0 on an
NVIDIA RTX~4070 Ti SUPER (16\,GB).
Language-field training uses render downscale 0.5; query decoding uses
threshold $\tau_q{=}0.55$ with open-dictionary negatives and multi-scale s/m/l decoding,
held fixed across compared open-vocabulary methods. Switching scenes resets
only the geometric map and embedding bank; SAM and CLIP weights remain
unchanged. All ablated hyperparameters are listed in Table~\ref{tab:ablation_neuro}.

\FloatBarrier
\section{Experiments}

\subsection{Experimental Setup}
\label{sec:experiments-setup}

\paragraph{Datasets.}
We evaluate in Habitat~\cite{savva2019habitat} on two photorealistic indoor benchmarks.
All methods in our comparisons receive the same ground-truth Habitat camera poses; none is
evaluated with estimated poses or SLAM localization noise.

\textbf{Replica}~\cite{straub2019replica}: eight scenes (\texttt{office0}--\texttt{office4},
\texttt{room0}--\texttt{room2}).
Replica follows the standard ActiveGS~\cite{jin2025activegs}/ActiveSGM~\cite{chen2025activesgm}
benchmark and therefore enables direct comparison with previous active mapping work.
Following ActiveGS~\cite{jin2025activegs}/ActiveSGM~\cite{chen2025activesgm}, each Replica
exploration is limited to 2000 steps.
RGB-D is captured at $680\times1200$.
Novel-view semantic evaluation uses held-out camera views that are not used for mapping.

\textbf{Matterport3D (MP3D)}~\cite{chang2017matterport3d}: five scenes from the benchmark
used by NARUTO~\cite{feng2024naruto}, ActiveGAMER~\cite{chen2025activegamer}, and
ActiveSGM~\cite{chen2025activesgm}. The scenes are \texttt{GdvgFV5R1Z5},
\texttt{gZ6f7yhEvPG}, \texttt{HxpKQynjfin}, \texttt{pLe4wQe7qrG}, and
\texttt{YmJkqBEsHnH}.
We use the same Habitat rendering settings as on Replica.
Novel-view semantic evaluation likewise uses held-out camera views that are not used for mapping.
We follow the MP3D protocol of ActiveSGM~\cite{chen2025activesgm},
ActiveGAMER~\cite{chen2025activegamer}, and NARUTO~\cite{feng2024naruto} using the same five
benchmark scenes and a 5000-step exploration budget.
No established passive semantic baseline exists for this protocol; therefore, passive
comparisons are reported only on Replica.

\paragraph{Baselines.}
\textbf{Passive} (Replica only, semantic evaluation): fixed NVS trajectory, SAM--CLIP
embedding extraction, and offline language-field training.
\textbf{Geometry-based}: geometry-only ActiveGS~\cite{jin2025activegs} planner with the same
SAM--CLIP and language-field pipeline.
\textbf{Geometry-based + CorrCLIP}: the same geometric planner with CorrCLIP applied to
accepted keyframe masks and embeddings; candidate views are still ranked only by geometry.
\textbf{ActiveSGM traj.\ + LangSplatV2}: we replay ActiveSGM's recorded pose sequence and
keyframe schedule on each scene, replacing OneFormer with our SAM--CLIP extraction and
LangSplatV2 language-field training.
This comparison keeps the open-vocabulary perception stack fixed, but does not isolate
trajectory quality completely because the replay contains 868 steps versus 956 for AOV-GS.
\textbf{ActiveSGM}~\cite{chen2025activesgm}: per-scene OneFormer fine-tuning on closed
Replica and Matterport3D label sets, run in our Habitat setup on both datasets; shown for
context only, not as an open-vocabulary baseline.
\textbf{AOV-GS (Ours)}: our semantic stage-0 planner without CorrCLIP.
\textbf{AOV-GS + CorrCLIP (Ours)}: the same semantic planner with CorrCLIP applied both to
candidate masks during re-ranking and to accepted keyframe masks used for the embedding bank.

\paragraph{Metrics.}
Primary: open-vocabulary language-field mIoU~\cite{everingham2010pascal} on text-query
pairs over held-out novel views.
For each dataset category present in a held-out view, its category name is used as the text
query, the dataset semantic annotation supplies the binary ground-truth mask, and IoU is
computed between that mask and the thresholded language-field prediction. We average over
all category--view pairs and then report the scene mean; for example, \texttt{office0}
contains 23 categories and 2477 pairs. The released evaluation configuration fixes the
held-out views, category names, negative prompts, and threshold for every compared method.
Secondary: novel-view PSNR~\cite{kelecs2021computation}, SSIM~\cite{wang2004image},
LPIPS~\cite{zhang2018unreasonable}, and depth L1 after map refinement.
For the reported MP3D reconstruction metrics, we evaluate the outputs at the end of stage~1
and final refinement and report the better result per scene.

\subsection{Open-Vocabulary Semantic Segmentation on Replica}

Table~\ref{tab:semantic_neuro} summarizes open-vocabulary language-field mIoU on eight
Replica scenes.
AOV-GS reaches 34.6\% average mIoU without CorrCLIP (+1.3 over Geometry-based,
+2.5 over Passive). With CorrCLIP, AOV-GS reaches 35.6\%, compared with 33.6\% for
Geometry-based + CorrCLIP.
ActiveSGM (78.8\% closed-set mIoU in our runs) uses per-scene OneFormer on fixed label sets
and is listed only as a supervised reference outside our open-vocabulary protocol.

\begin{table}[t]
    \centering
    \caption{Open-vocabulary language-field mIoU (\%) on Replica novel views.
    Gray rows: closed-set OneFormer runs in our setup (context only, not open-vocabulary baselines).}
    \label{tab:semantic_neuro}
    \setlength{\tabcolsep}{3pt}
    \adjustbox{max width=\textwidth}{%
    \begin{tabular}{lrrrrrrrrr}
        \toprule
        Method & Avg. & Of0 & Of1 & Of2 & Of3 & Of4 & R0 & R1 & R2 \\
        \midrule
        \rowcolor{gray!20}
        ActiveSGM (closed-set) & 78.8 & 74.5 & 74.8 & 84.9 & 85.0 & 76.2 & 70.1 & 82.2 & 82.5 \\
        ActiveSGM trajectory + LangSplatV2 & 28.2 & 32.5 & 27.4 & 37.1 & 27.5 & 8.9 & 23.1 & 40.2 & 28.7 \\
        \midrule
        Passive            & 32.1 & 31.3 & 22.8 & 29.5 & 31.1 & 36.4 & 35.3 & 39.1 & \textbf{30.9} \\
        Geometry-based         & 33.3 & 27.3 & \textbf{31.3} & 32.6 & 34.1 & 35.3 & 36.4 & 39.7 & 29.3 \\
        Geometry-based + CorrCLIP & 33.6 & 30.8 & 29.7 & 34.1 & 33.7 & 35.2 & \textbf{36.5} & 39.0 & 29.5 \\
        AOV-GS (Ours)      & 34.6 & \textbf{34.4} & 30.1 & 28.7 & \textbf{39.3} & \textbf{38.7} & 34.7 & \textbf{42.8} & 27.9 \\
        AOV-GS + CorrCLIP (Ours) & \textbf{35.6} & 33.6 & 29.8 & \textbf{37.3} & 38.5 & 38.3 & 36.4 & 41.7 & 29.4 \\
        \bottomrule
    \end{tabular}}
\end{table}

To better understand the heterogeneous per-scene results, we analyzed scene size, object
statistics, lighting conditions, and semantic evaluation subsets.
No consistent correlation was observed with scene area or object count alone.
Instead, improvements were primarily determined by the reliability of semantic observations.
Large, well-segmented objects produced stable SAM masks and CLIP embeddings, whereas adverse
visual conditions (including low illumination, overexposure, and cluttered scenes containing
many small objects) reduced semantic feature quality and limited the benefit of semantic
planning.
The CorrCLIP improvement on \texttt{office2} (+8.6~pp) suggests that semantic feature
quality is an important bottleneck. CorrCLIP is applied both to candidate masks used for
semantic re-ranking and to accepted keyframe masks used to populate the embedding bank, so
this result combines planner and representation effects.

\paragraph{Failure modes.}
The four Replica regressions have different signatures (Table~\ref{tab:semantic_neuro}).
\texttt{office1} is the darkest Replica scene in our image-statistics analysis (mean RGB
intensity 44/255; 5th percentile 4/255), which may reduce the reliability of SAM masks and
CLIP crops; AOV-GS is 1.2~pp below Geometry-based. We retain the benchmark's fixed rendering
conditions rather than tuning illumination per method. On \texttt{office2}, AOV-GS is
3.9~pp below Geometry-based, while CorrCLIP recovers 8.6~pp; this is consistent with mask
fragmentation being an important factor. \texttt{room0} shows a category trade-off rather
than a uniform failure: relative to Passive, AOV-GS improves floor IoU from 17.4\% to 57.2\%
and chair from 75.2\% to 82.8\%, while its full-category mean decreases from 35.3\% to 34.7\%
(Table~\ref{tab:perclass_replica}). On \texttt{room2}, the largest displayed degradation is
chair IoU, from 76.3\% to 32.8\%, although floor improves from 7.0\% to 57.6\%
(Table~\ref{tab:perclass_replica}); this points to sensitivity to which object views are
collected rather than a uniform loss of reconstruction quality.

\begin{figure}[ht]
    \centering
    \methodfig{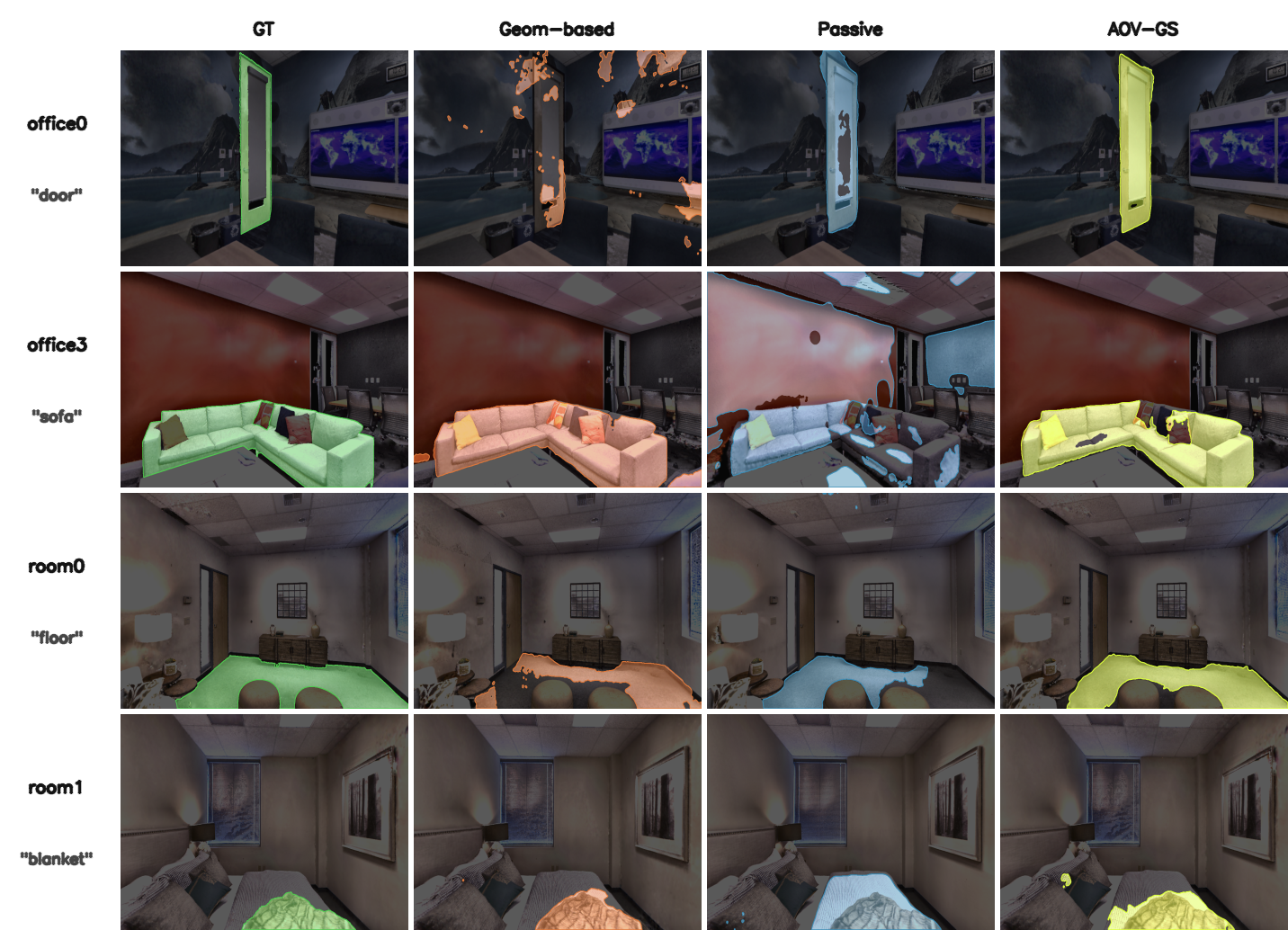}{Cross-scene language-field mask comparison}
    \caption{Language-field masks on novel views for queries door, sofa, floor, and blanket.
    AOV-GS improves large-object coverage (\texttt{office3}/sofa, \texttt{room0}/floor);
    narrow objects such as the \texttt{office0} door remain difficult for all methods.}
    \label{fig:mask_neuro}
\end{figure}

Figure~\ref{fig:mask_neuro} qualitatively illustrates representative improvements on large
structural objects together with typical failure cases on narrow or visually ambiguous
categories.

\begin{figure}[ht]
    \centering
    \methodfig{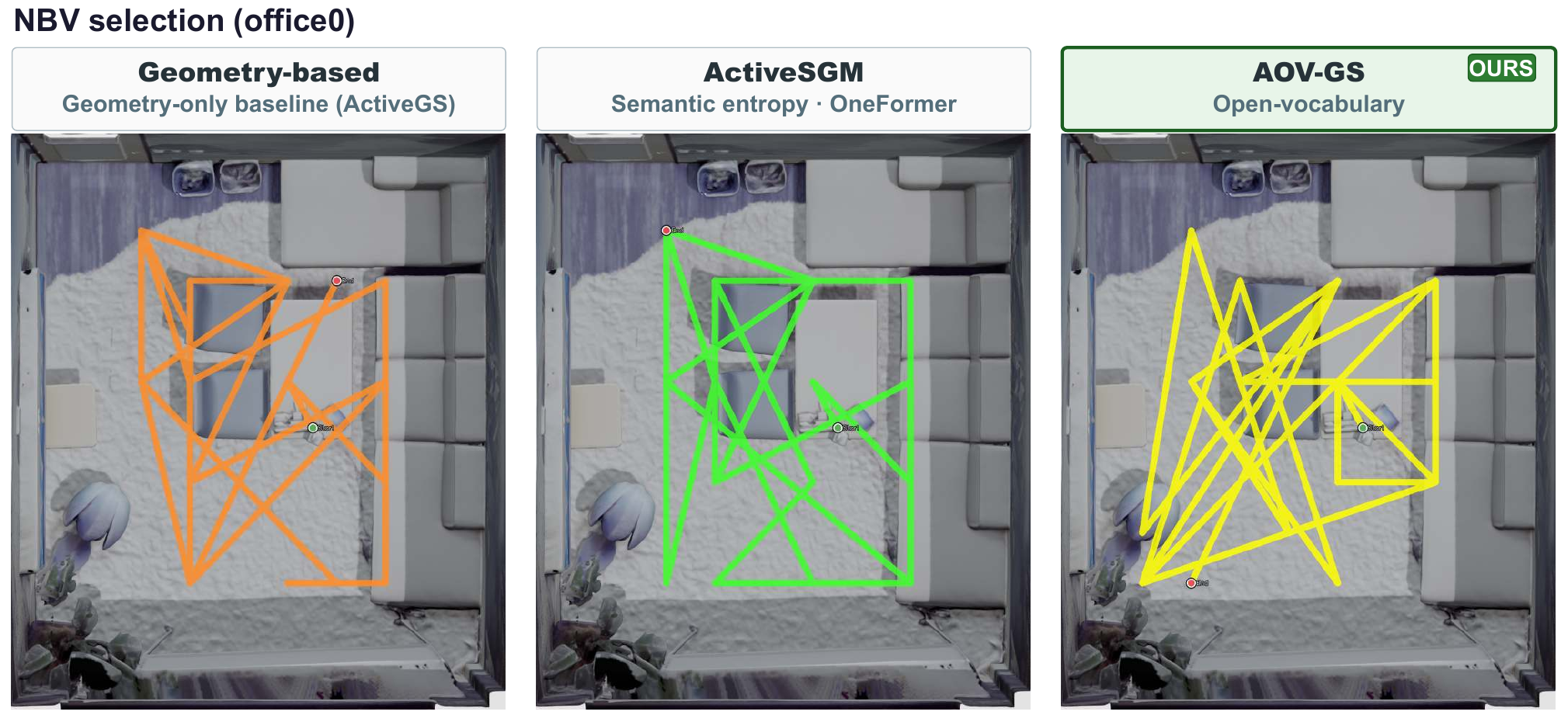}{Exploration trajectories on office0}
    \caption{Exploration routes on \texttt{office0}: AOV-GS covers the central workspace
    more thoroughly than the geometry-only planner.}
    \label{fig:fusion_neuro}
\end{figure}

\paragraph{When semantic exploration helps.}
Semantic exploration is most beneficial when semantically distinctive objects occupy the
central workspace and remain visible from multiple viewpoints.
On \texttt{office0}, the geometry-only planner favours peripheral unseen regions over the
furniture-dense core (Fig.~\ref{fig:fusion_neuro}), yielding +7.1~pp mIoU over the
geometry-only baseline (Table~\ref{tab:semantic_neuro}) and strong \emph{sofa}/\emph{table}
coverage for AOV-GS (87.3\%/84.3\%).
Similar behaviour appears on \texttt{office3} and \texttt{office4} (+4.4 and +3.0~pp with
CorrCLIP).
In the bedroom scene \texttt{room1}, discriminative bed-region embeddings give the best
open-vocabulary result (42.8\%).
Table~\ref{tab:perclass_replica} lists per-category IoU for structural classes.

\begin{table}[t]
    \centering
    \caption{Selected per-category language-field IoU (\%) on Replica scenes: AOV-GS vs.\ Passive.
    Higher value in bold per category; ``---'' marks categories absent from a scene.
    The Mean row is computed over the full evaluation category set, not only the displayed rows.}
    \label{tab:perclass_replica}
    \setlength{\tabcolsep}{1.5pt}
    \adjustbox{max width=\textwidth}{%
    \begin{tabular}{lcccccccccccccccc}
        \toprule
        & \multicolumn{2}{c}{Of0} & \multicolumn{2}{c}{Of1} & \multicolumn{2}{c}{Of2}
        & \multicolumn{2}{c}{Of3} & \multicolumn{2}{c}{Of4}
        & \multicolumn{2}{c}{R0} & \multicolumn{2}{c}{R1} & \multicolumn{2}{c}{R2} \\
        \cmidrule(lr){2-3}\cmidrule(lr){4-5}\cmidrule(lr){6-7}\cmidrule(lr){8-9}\cmidrule(lr){10-11}
        \cmidrule(lr){12-13}\cmidrule(lr){14-15}\cmidrule(lr){16-17}
        Category & Ours & Pass. & Ours & Pass. & Ours & Pass. & Ours & Pass. & Ours & Pass.
        & Ours & Pass. & Ours & Pass. & Ours & Pass. \\
        \midrule
        bed     & --- & --- & --- & --- & --- & --- & --- & --- & --- & ---
        & --- & --- & \textbf{26.5} & 22.0 & --- & --- \\
        sofa    & \textbf{87.3} & 70.6 & --- & --- & 41.3 & \textbf{54.8} & \textbf{66.6} & 31.7 & --- & ---
        & 29.5 & \textbf{32.7} & --- & --- & --- & --- \\
        table   & \textbf{84.3} & 66.8 & --- & --- & 45.6 & \textbf{57.5} & \textbf{77.0} & 62.5 & 0.3 & \textbf{35.5}
        & 45.3 & \textbf{59.1} & --- & --- & \textbf{81.3} & 76.9 \\
        chair   & 36.4 & \textbf{73.8} & \textbf{56.3} & 29.1 & \textbf{48.2} & 33.4 & 73.1 & \textbf{75.7} & \textbf{28.6} & 27.0
        & \textbf{82.8} & 75.2 & --- & --- & 32.8 & \textbf{76.3} \\
        wall    & \textbf{39.7} & 33.6 & \textbf{22.0} & 14.2 & 59.0 & \textbf{70.6} & \textbf{57.5} & 47.6 & \textbf{57.5} & 52.8
        & \textbf{15.8} & 13.6 &  9.1 & \textbf{24.6} & 12.0 & \textbf{43.4} \\
        floor   &  4.9 & \textbf{8.7}  & \textbf{22.6} & 2.1 &  9.3 & \textbf{15.3} & \textbf{45.9} & 21.0 & \textbf{41.6} & 20.1
        & \textbf{57.2} & 17.4 & \textbf{64.5} & 55.1 & \textbf{57.6} &  7.0 \\
        ceiling & \textbf{50.3} & 17.5 & \textbf{39.9} & 0.1 & 39.9 & \textbf{47.0} & 59.5 & \textbf{73.8} & \textbf{72.1} & 67.6
        & \textbf{68.1} & 61.5 & \textbf{67.5} & 62.9 & \textbf{45.0} & 44.6 \\
        pillow  & --- & --- & --- & --- & --- & --- & --- & --- & --- & ---
        & --- & --- & \textbf{80.9} & 79.2 & --- & --- \\
        \midrule
        Mean    & \textbf{34.4} & 31.3 & \textbf{30.1} & 22.8 & 28.7 & \textbf{29.5} & \textbf{39.3} & 31.1 & \textbf{38.7} & 36.4
        & 34.7 & \textbf{35.3} & \textbf{42.8} & 39.1 & 27.9 & \textbf{30.9} \\
        \bottomrule
    \end{tabular}}
\end{table}

\paragraph{Trajectory transfer.}
The ActiveSGM traj.\ + LangSplatV2 row in Table~\ref{tab:semantic_neuro} averages 28.2\% mIoU,
below both Geometry-based (33.3\%) and AOV-GS (34.6\%).
A trajectory optimized for OneFormer supervision does not benefit a SAM--CLIP language field;
the largest degradation occurs on \texttt{office4} (8.9\%), while \texttt{room1} remains
high (40.2\%) because furniture layout still yields useful views regardless of planner objective.
The 6.4~pp gap between this transferred trajectory and AOV-GS shows that viewpoint selection
must be matched to the semantic representation used for training and querying the map.

\FloatBarrier
\subsection{RGB-D Reconstruction on Replica}

A practical semantic planner should preserve reconstruction quality while improving semantic
coverage.
Table~\ref{tab:rgbd_neuro} reports eight-scene averages from a single run per method.
AOV-GS matches the Geometry-based planner on PSNR (31.19 vs.\ 31.47\,dB), SSIM, and LPIPS,
while improving depth L1 (1.18 vs.\ 1.26\,cm).

To assess stability, we repeat Geometry-based and AOV-GS on all eight scenes with five
independent seeds (0--4).
Table~\ref{tab:multiseed_neuro} summarizes pooled rendering metrics and \texttt{office0}
language-field mIoU.
Cross-seed PSNR, SSIM, and LPIPS have similar descriptive means and spreads. On \texttt{office0}, semantic
mIoU is $33.6{\pm}1.1\%$ for AOV-GS versus $28.8{\pm}1.4\%$ for Geometry-based.

\begin{table}[t]
    \centering
    \caption{Novel-view rendering after refinement on Replica (eight-scene average,
    single run).}
    \label{tab:rgbd_neuro}
    \begin{tabular}{lccccc}
        \toprule
        Method & Steps$\downarrow$ & PSNR$\uparrow$ & SSIM$\uparrow$ & LPIPS$\downarrow$ & L1 (cm)$\downarrow$ \\
        \midrule
        Passive            & 1999 & 27.45 & 0.949 & 0.149 & 6.98 \\
        Geometry-based         & 914  & 31.47 & 0.964 & 0.131 & 1.26 \\
        ActiveSGM          & \textbf{868}  & \textbf{32.11} & \textbf{0.970} & \textbf{0.110} & 1.23 \\
        AOV-GS (Ours)      & 956  & 31.19 & 0.964 & 0.123 & 1.18 \\
        AOV-GS + CorrCLIP (Ours) & 952  & 31.56 & 0.964 & 0.128 & \textbf{1.14} \\
        \bottomrule
    \end{tabular}
\end{table}

\begin{table}[t]
    \centering
    \caption{Five-seed stability on Replica (seeds 0--4).
    Rendering: eight-scene average; mIoU: \texttt{office0} only.}
    \label{tab:multiseed_neuro}
    \begin{tabular}{lcccc}
        \toprule
        Method & PSNR$\uparrow$ & SSIM$\uparrow$ & LPIPS$\downarrow$ & mIoU (\%) \\
        \midrule
        Geometry-based & $\textbf{31.62}{\pm}2.63$ & $\textbf{0.965}{\pm}0.014$ & $0.128{\pm}0.033$ & $28.8{\pm}1.4$ \\
        AOV-GS (Ours)  & $31.53{\pm}2.44$ & $\textbf{0.965}{\pm}0.014$ & $\textbf{0.126}{\pm}0.027$ & $\textbf{33.6}{\pm}1.1$ \\
        \bottomrule
    \end{tabular}
\end{table}

\subsection{Computational Cost}

Stage-0 semantic scoring applies only to top-$K$ candidates (${\approx}8$\,s per view on
\texttt{office0} on an NVIDIA RTX~4070 Ti SUPER); stage~1+ adds no semantic overhead.
A complete \texttt{office0} run takes 52:32, 61:23, and 70:41 for $K{=}4$, 8, and 16,
respectively, with peak GPU memory of 12.59, 12.92, and 13.80\,GB. Because candidate views
are scored sequentially, increasing $K$ affects runtime more strongly than memory.
Offline language-field training peaks at 12.6\,GB with the default 0.5 downscale and ranges
from 9.9 to 29.1\,GB across the evaluated render-downscale settings;
runtime increases with the number of rendered pixels. Multi-scale text-query decoding takes
1.41\,s per query--view pair and peaks at 4.2\,GB.
On Matterport3D, both open-vocabulary planners fit in 16\,GB on this hardware, whereas
ActiveSGM in our setup required ${\approx}$36--48\,GB due to the OneFormer module.

\subsection{Evaluation on Matterport3D}

We apply the same open-vocabulary pipeline to the five MP3D scenes defined in
Sec.~\ref{sec:experiments-setup}.
Compared with Replica, MP3D contains larger layouts and substantially longer exploration
trajectories, reducing the relative influence of the semantic stage-0 planner.
Table~\ref{tab:mp3d_neuro} reports language-field mIoU on MP3D novel views.
AOV-GS improves the average from 21.3\% to 23.1\% (+1.8~pp) over the Geometry-based
planner, with the largest absolute gains on \texttt{GdvgFV5R1Z5} (+3.5~pp),
\texttt{gZ6f7yhEvPG} (+2.7~pp), and \texttt{pLe4wQe7qrG} (+1.9~pp).
The improvement on \texttt{HxpKQynjfin} (+0.7~pp) supports that semantic novelty remains
informative in larger layouts.
On MP3D, exploration trajectories are longer (mean 3472 vs.\ 3496 steps), but semantic
stage-0 re-ranking still shifts viewpoint selection toward CLIP-embedding novelty.

\begin{table}[t]
    \centering
    \caption{Open-vocabulary language-field mIoU (\%) on Matterport3D novel views
    (five scenes). Gray rows: closed-set OneFormer runs in our setup (context only).}
    \label{tab:mp3d_neuro}
    \setlength{\tabcolsep}{4pt}
    \adjustbox{max width=\textwidth}{%
    \begin{tabular}{@{}l r r r r r r@{}}
        \toprule
        Method & Avg. & HxpK & Gdvg & gZ6f & YmJk & pLe4 \\
        \midrule
        \rowcolor{gray!20}
        ActiveSGM (closed-set)
            & 61.5 & 51.2 & 56.3 & 69.8 & 66.1 & 64.0 \\
        ActiveSGM trajectory + LangSplatV2
            & 18.1 & 25.2 & 16.9 & 19.7 & 13.8 & 15.1 \\
        \midrule
        Geometry-based & 21.3 & 29.6 & 19.9 & 23.2 & 16.2 & 17.8 \\
        AOV-GS (Ours)  & 23.1 & 30.3 & 23.4 & 25.9 & 16.2 & 19.7 \\
        Geometry-based + CorrCLIP
            & 21.8 & 29.7 & 20.1 & 23.6 & 17.6 & 18.0 \\
        AOV-GS + CorrCLIP (Ours)
            & \textbf{24.0} & \textbf{30.5} & \textbf{23.5} & \textbf{26.8} & \textbf{18.1} & \textbf{21.1} \\
        \bottomrule
    \end{tabular}
    }
\end{table}

The CorrCLIP gain is smaller for Geometry-based because its trajectory is fixed by
the geometric planner: CorrCLIP can improve only the masks and embeddings extracted from
accepted keyframes. In AOV-GS, CorrCLIP additionally changes the candidate-mask embeddings
used for semantic re-ranking and can therefore alter the acquired trajectory. CorrCLIP is
not uniformly beneficial, however: merging adjacent regions or suppressing small and
overlapping masks can remove useful object evidence, so its effect depends on the scene.

With CorrCLIP, AOV-GS reaches 24.0\% average mIoU, compared with 21.8\% for the matched
Geometry-based + CorrCLIP baseline. The resulting +2.2~pp difference isolates the benefit
of semantic view selection under the same CorrCLIP-enhanced perception stack.

\begin{figure}[t]
    \centering
    \IfFileExists{figures/mask_compare_mp3d_selected.png}{%
        \adjustbox{max width=\linewidth,max height=0.35\textheight,keepaspectratio}{%
            \includegraphics{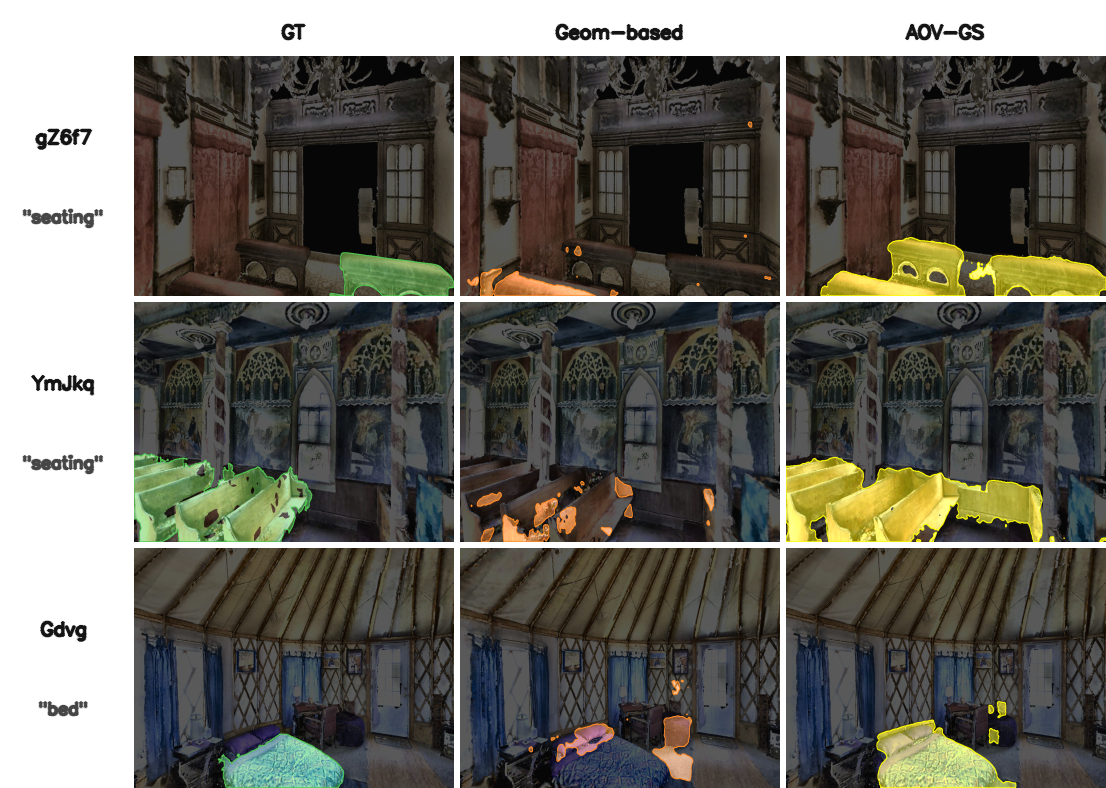}}%
    }{%
        \fbox{\parbox{0.98\linewidth}{\centering\vspace{1.2cm}%
        \textbf{MP3D language-field mask comparison}\vspace{1.2cm}}}%
    }
    \caption{Language-field masks on MP3D novel views for \texttt{gZ6f7}/seating,
    \texttt{HxpKQ}/chair, \texttt{YmJkq}/seating, and \texttt{Gdvg}/bed (GT in green;
    Geometry-based in orange; AOV-GS in yellow).}
    \label{fig:mask_mp3d}
\end{figure}

Although absolute mIoU values are lower on Matterport3D, AOV-GS retains a +1.8~pp average
advantage over the geometry-only planner (Fig.~\ref{fig:mask_mp3d}).

\FloatBarrier
Novel-view rendering on MP3D is summarized in Table~\ref{tab:mp3d_rgbd_neuro}.
AOV-GS improves novel-view rendering quality over the geometry-only planner on PSNR, SSIM,
and LPIPS; depth L1 is the only metric slightly above the geometry baseline
(2.83 vs.\ 2.80\,cm).
The MP3D results therefore retain the semantic gain without reducing RGB rendering quality.
\begin{table}[t]
    \centering
    \caption{Novel-view rendering on Matterport3D (five-scene average, single run); each
    scene uses the better offline evaluation of the stage-1 and final-refinement outputs.}
    \label{tab:mp3d_rgbd_neuro}
    \begin{tabular}{lccccc}
        \toprule
        Method & Steps$\downarrow$ & PSNR$\uparrow$ & SSIM$\uparrow$ & LPIPS$\downarrow$ & L1 (cm)$\downarrow$ \\
        \midrule
        Geometry-based              & 3496 & 26.19 & 0.928 & 0.233 & 2.80 \\
        ActiveSGM                   & 3821 & 26.08 & 0.917 & 0.259 & 3.76 \\
        AOV-GS (Ours)               & 3472 & 26.25 & 0.931 & \textbf{0.232} & 2.83 \\
        AOV-GS + CorrCLIP (Ours)    & \textbf{3454} & \textbf{26.27} & \textbf{0.932} & 0.233 & \textbf{2.79} \\
        \bottomrule
    \end{tabular}
\end{table}

\subsection{Ablation Studies}

Table~\ref{tab:ablation_neuro} summarizes single-run ablations on \texttt{office0} over 23 categories
(2477 query pairs), varying one factor at a time while keeping the default configuration
elsewhere ($K{=}8$, render downscale~$0.5$, threshold~$0.55$, LangSplatV2
``other'' negatives).
Increasing top-$K$ from 4 to 8 improves mIoU from 33.2\% to 34.4\%; $K{=}16$ does not
improve quality and increases semantic-scoring time because twice as many candidates are
processed relative to $K{=}8$.
Render downscale~$0.5$ balances mIoU and offline training cost; full resolution
(downscale~$1.0$) overfits CLIP feature noise (26.2\%), while downscale~$0.25$ reaches
32.0\% mIoU with lower offline memory (9.9\,GB) and downscale~$0.5$ reaches the best mIoU 34.4\%.
LangSplatV2 ``other'' negatives with $\tau_q{=}0.55$ outperform alternatives; query-only
decoding reaches 25.1\% mIoU.

\begin{table}[t]
    \centering
    \caption{Ablation on \texttt{office0}. GPU memory denotes peak allocated memory.}
    \label{tab:ablation_neuro}
    \setlength{\tabcolsep}{3pt}
    \adjustbox{max width=\textwidth}{%
    \begin{tabular}{@{}l l r c @{}}
        \toprule
        Factor & Setting & mIoU$\uparrow$ & Peak GPU memory \\
        \midrule
        \multirow{3}{*}{Top-$K$}
            & $K{=}4$ & 33.2 & 12.59\,GB \\
            & $K{=}8$ (default) & \textbf{34.4} & 12.92\,GB \\
            & $K{=}16$ & 34.4 & 13.80\,GB \\
        \midrule
        \multirow{3}{*}{Downscale}
            & $0.25$ & 32.0 & 9.9\,GB \\
            & $0.5$ (default) & \textbf{34.4} & 12.6\,GB \\
            & $1.0$ (full) & 26.2 & 29.1\,GB \\
        \midrule
        \multirow{3}{*}{\shortstack[l]{Negative\\prompts ($\tau_q{=}0.55$)}}
            & none & 25.1 & 4.2\,GB \\
            & 1 class vs.\ all others & 17.7 & 4.2\,GB \\
            & 1 class vs.\ ``other'' (LangSplatV2) & \textbf{34.4} & 4.2\,GB \\
        \midrule
        \multirow{3}{*}{\shortstack[l]{Threshold\\(``other'' negs.)}}
            & $\tau_q{=}0.50$ & 25.0 & 4.2\,GB \\
            & $\tau_q{=}0.55$ (default) & \textbf{34.4} & 4.2\,GB \\
            & $\tau_q{=}0.60$ & 33.9 & 4.2\,GB \\
        \bottomrule
    \end{tabular}}
\end{table}

\FloatBarrier
\section{Conclusion}

We presented AOV-GS, an active open-vocabulary semantic mapping pipeline that guides
exploration with CLIP mask-embedding novelty from frozen SAM and CLIP and trains a
scene-specific LangSplatV2 language field on the collected embeddings for text-query
segmentation on novel views.
Across Replica and Matterport3D, the resulting maps improve average open-vocabulary
language-field mIoU over geometry-only exploration while preserving comparable novel-view
reconstruction quality. Replica gains vary by scene, while the five-seed \texttt{office0}
experiment shows a consistent advantage over geometry-only exploration.

Experiments further show that these gains depend on alignment between exploration and the
downstream semantic representation: trajectories optimized for a closed-set OneFormer
pipeline do not transfer to a SAM--CLIP language field, whereas semantic novelty
co-designed with that representation improves the average on both benchmarks.
The evaluation uses ground-truth camera poses, matching the simulator protocol used by the
compared active-mapping methods; robustness to SLAM drift and real sensor noise is outside
the scope of this simulator study.

\subsubsection*{Disclosure of Interests.}
The authors have no competing interests to declare that are relevant to the content of this article.

\bibliographystyle{splncs04}
\bibliography{references}

\end{document}